\pdfoutput=1

\documentclass[11pt]{article}

\usepackage[preprint]{acl}
\usepackage{booktabs}

\usepackage{times}
\usepackage{latexsym}

\usepackage{enumitem} 
\usepackage{parskip} 

\usepackage[T1]{fontenc}

\usepackage[utf8]{inputenc}

\usepackage{microtype}
\usepackage{hyperref}
\usepackage{inconsolata}

\usepackage{graphicx}

\title{QUB-Cirdan at ``Discharge Me!'': Zero shot discharge letter generation by open-source LLM}

\iftrue 

 \author{
  \textbf{Rui Guo\textsuperscript{1,2}} \thanks{rui.guo@cirdan.com} \ \ \ \ 
  \textbf{Greg Farnan\textsuperscript{2}} \thanks{greg.farnan@cirdan.com} \ \ \ \  
  \textbf{Niall McLaughlin\textsuperscript{1}} \thanks{n.mclaughlin@qub.ac.uk} \ \ \ \  
  \textbf{Barry Devereux\textsuperscript{1}} \thanks{b.devereux@qub.ac.uk}
\\
  \textsuperscript{1} Queen's University Belfast \\
  \textsuperscript{2} Cirdan \\
}

\fi 

\begin{document}
\maketitle
\begin{abstract}
The BioNLP ACL'24 Shared Task on Streamlining Discharge Documentation aims to reduce the administrative burden on clinicians by automating the creation of critical sections of patient discharge letters. This paper presents our approach using the Llama3 8B quantized model to generate the ``Brief Hospital Course'' and ``Discharge Instructions'' sections. We employ a zero-shot method combined with Retrieval-Augmented Generation (RAG) to produce concise, contextually accurate summaries. Our contributions include the development of a curated template-based approach to ensure reliability and consistency, as well as the integration of RAG for word count prediction. We also describe several unsuccessful experiments to provide insights into our pathway for the competition. Our results demonstrate the effectiveness and efficiency of our approach, achieving high scores across multiple evaluation metrics.

\end{abstract}

\section{Introduction}

The BioNLP ACL'24 Shared Task, ``Discharge Me!'' on Codabench \citep{xu-etal-2024-overview}, focuses on automating the creation of two crucial sections of patient discharge letters: ``Brief Hospital Course'' (BHC) and ``Discharge Instructions'' (DI). This initiative arises in response to significant time burdens on clinicians, highlighted by surveys of U.S. physicians. One study found that physicians spend twice as much time on Electronic Health Records (EHR) compared to direct patient interactions during clinical hours \citep{sinsky2016allocation}. Another survey involving 1,524 physicians revealed an average of 1.84 hours spent on EHR documentation outside office hours. Automating the generation of BHC and DI aims to significantly reduce the clerical load on healthcare providers, thereby improving patient service quality and potentially mitigating clinician burnout.

A discharge letter, or a discharge summary, is a critical document summarizing a patient's hospital visit from admission to discharge, serving as a bridge between hospital care and follow-up with outpatient providers. Among its several sections, the ``Brief Hospital Course'' outlines the patient's treatment and progress during the hospital stay, typically using clinical jargon best understood by healthcare professionals. Conversely, the ``Discharge Instructions'' are designed to guide patients and their caregivers once they leave the hospital, using layman’s language to clearly explain follow-up care, medication regimens, and lifestyle recommendations.

Large Language Models (LLMs) offer a promising solution for automating medical documentation due to their ability to understand and generate human-like text \citep{singhal2023large, zhanghuatuo}. Unlike traditional extractive summarization \cite{el2021automatic}, which predominantly involves concatenating snippets from existing texts, LLMs can enhance summarization by integrating both extractive and abstractive techniques. This has been applied to progress note summarization \citep{gao2022summarizing, liu2023deakinnlp}, similar to this Codabench challenge. With both proprietary LLMs such as ChatGPT \citep{openai2023chatgpt} and open-source LLMs such as Llama3 \citep{llama3modelcard}, the potential for creating accessible medical summaries is significant. 

In this challenge, we propose a zero-shot approach utilizing the Llama3 8B quantized model, which is optimized for low computing resource usage without fine-tuning, and the result is in the top 10 in the final benchmark assessment. Our key contributions are:

\begin{itemize}
    \item Crafting specialized templates for the ``Brief Hospital Course'' and ``Discharge Instructions'' sections, with carefully designed prompts to ensure the generated text is medically reliable and stylistically consistent with the training dataset.
    \item Exploring various methods to estimate the total word count for the target sections, including:
    \begin{itemize}
        \item Fitting a statistical distribution
        \item Employing a random forest classifier
        \item Implementing a context-based retrieval system
    \end{itemize}
    \item Conducting all experiments using a T4 GPU, demonstrating that our approach is computationally efficient.
\end{itemize}

\section{Related Work}
The application of foundation models, pre-trained on billions of tokens from diverse data sources, is increasingly prevalent in healthcare \citep{he2024foundation}. These models are pivotal in various domains, such as diagnosis generation \citep{gao2023leveraging} and medical image analysis \citep{zhang2024sam}. Within clinical text processing, large language models (LLMs) are employed for tasks including summarization \citep{van2023clinical, gao2023overview} and answering medical questions \citep{singhal2023towards}. Specifically, the ``Discharge me!'' challenge involves condensing extensive medical records into succinct discharge letters while retaining all critical information, making LLMs suited for this task.

Participants in the BioNLP 2023 Workshop's Problem List Summarization task often utilized T5 \citep{raffel2020exploring} or BART \citep{lewis2019bart} models, enhancing these backbones either by further training on clinical texts or fine-tuning for specific clinical tasks \citep{gao2023overview}. This further pre-training introduces medical knowledge not originally present in the LLM while fine-tuning adapts the model to produce outputs in the correct format for the target task.

Several studies such as BioMistral \citep{labrak2024biomistral} and PMC-LLaMA \citep{wu2024pmc} have adapted open-source LLMs by applying pre-training and fine-tuning sequentially. Conversely, Med-PaLM \citep{singhal2023large} bypasses additional pre-training, relying solely on fine-tuning from a vast pre-trained dataset. On a different note, BioMedLM \citep{bolton2024biomedlm} focuses exclusively on medical texts, resulting in a smaller model but still competes effectively with models trained on larger, more general datasets.

Pre-training and fine-tuning LLMs require GPUs with significant memory capacities (often exceeding 16GB). Fine-tuning can take several days, even using Parameter-Efficient Fine Tuning (PEFT) methods like LoRA \citep{hu2021lora}. However, modern LLMs can exhibit strong performance without additional fine-tuning if provided with the appropriate context and instructions. For instance, Almanac \citep{zakka2024almanac} enhances its output by retrieving clinical question-related knowledge from curated sources, a technique known as Retrieval-augmented Generation (RAG) \citep{gao2023retrieval}. Additionally, Medagents \citep{tang2023medagents} demonstrates that a zero-shot method, which deconstructs the question into distinct steps and assigns specific prompts and roles to the LLM for each stage, can achieve competitive results compared to more traditional few-shot approaches.

\section{Methods}

\begin{figure*}[htpb]
  \includegraphics[width=1\linewidth]{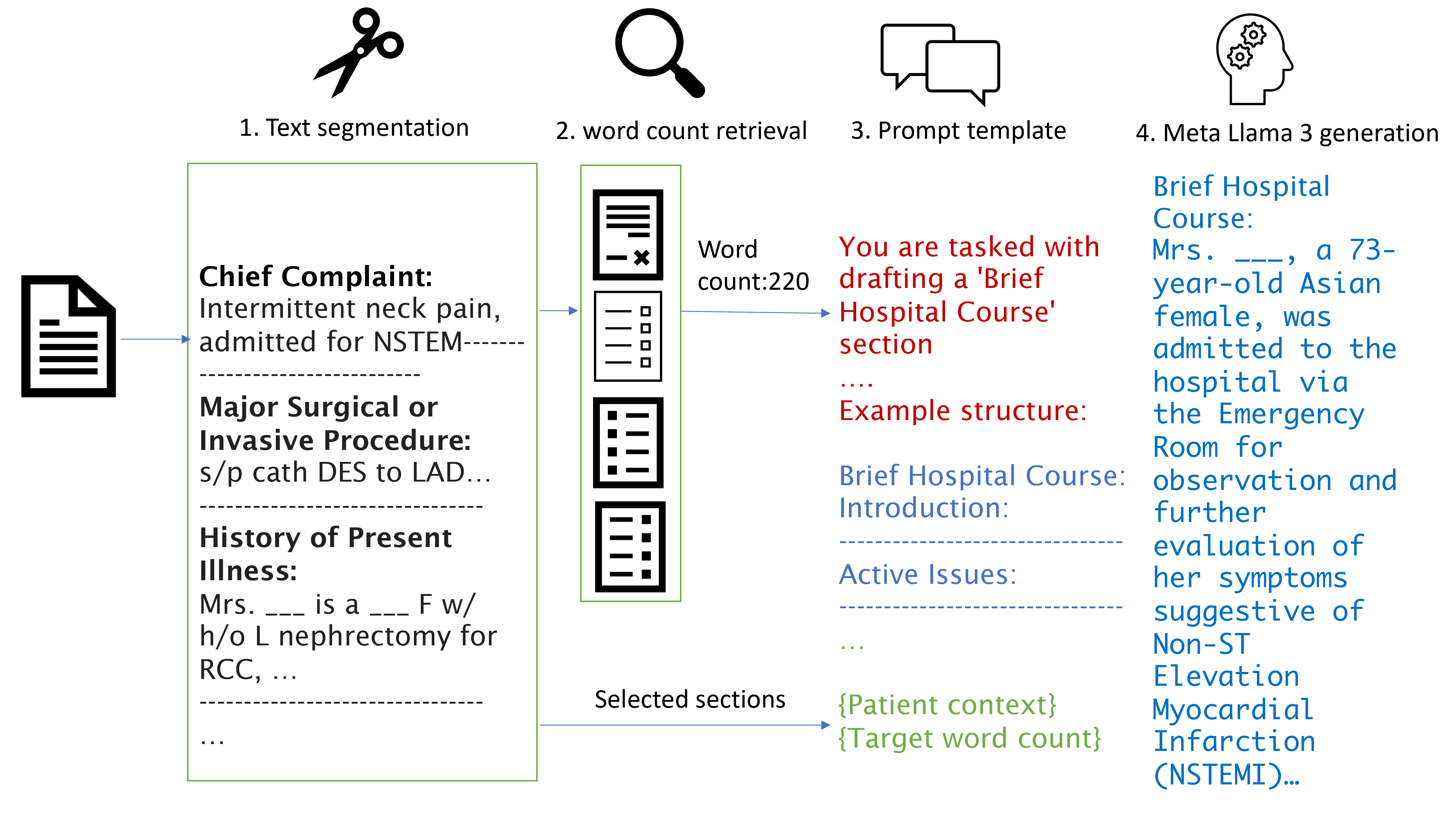}
  \caption{Overview of our solution. The figure illustrates our four-step approach: (1) Text Segmentation: splitting the discharge letter into sections such as ``Chief Complaint'' and ``Brief Hospital Course''; (2) Retrieval-Augmented Generation (RAG): retrieving similar patient sections to determine word count; (3) Template and Prompt Design: providing structured templates and prompts to Llama3 with patient context and target word count; (4) Text Generation: generating the final output using Llama3. }
  \label{fig:overview}
\end{figure*}

In this section, we introduce our zero-shot template-based approach, combined with RAG, to determine the target word count, which is both effective and resource-friendly. We adopted the Llama3 8B model with 8-bit quantization as the open-source model for this challenge. Figure \ref{fig:overview} illustrates our approach:

\begin{enumerate}
    \item Splitting the full discharge letter into different segments, such as ``Chief Complaint'' and ``Brief Hospital Course''. This allows us to selectively use relevant sections and discard or truncate those too lengthy to process.
    \item Employing Retrieval-Augmented Generation (RAG) to find the most similar patient's target section, using that section's word count as the target for generation. Generating a similar word count to the target can help maintain the generated summaries' completeness and increase evaluation metrics such as BLEU, ROUGE, and METEOR.
    \item Providing the target section's structure template and prompt to Llama3 along with the patient's context and target word count.
    \item Generating the result by Llama3 8B quantized model.
\end{enumerate}

While GPT-4/3.5 models generally outperform open-source models such as Llama2 in understanding EHR data \citep{liu2024evaluation}, the rules of this challenge discourage the use of proprietary model APIs (e.g., OpenAI's GPT-4). Consequently, we resorted to the state-of-the-art (SOTA) open-source model, Llama3 \citep{llama3modelcard}. Our approach leverages the full text from the ``text'' field in the provided discharge.csv file, alongside aggregated fields from other MIMIC-IV tables, including patient information, diagnoses, and transfer history.
We meticulously curated a template for each target section and designed prompts to guide the LLM in generating the required sections. In addition to our final approach, we documented several other zero-shot methods for target section generation and various approaches to predict the target section's word count. However, these were not adopted in our final solution.

\subsection{Dataset Exploration}

The dataset for this challenge is derived from MIMIC-IV's submodules, MIMIC-IV-Note \citep{johnson2023note} and MIMIC-IV-ED \citep{johnson2023ed}. All patients have visited the Emergency Department (ED), and the final target sections, ``Brief Hospital Course'' and ``Discharge Instructions'', are extracted from their discharge letters. Since patients can be admitted to the hospital after their initial ED visit, we also explored other tables from the MIMIC-IV hosp and ICU modules \citep{johnson2023mimiciv} to provide a comprehensive view of the patient's hospital stay beyond the ED information.

Due to limited context length, we could not simply pass all available information into the LLM. Therefore, we ranked all sections of the discharge letter to select a subset of the information. We segmented the discharge letter's ``text'' column from discharge.csv using regex and a template of keywords for different sections, as shown in the Section column of Table \ref{tab:ranking}. Besides the information from the ``text'' column, we aggregated ``Patient Admissions'' information, including gender, race, age (calculated), ``Diagnoses'' (throughout the patient stay), and ``Transfer Summary'' from other MIMIC-IV tables.  Since we compiled the patient's diagnoses and transfer summary for the entire hospital stay using other MIMIC-IV tables rather than just the Emergency Department (ED) stay, we did not use the tables in the ED module, such as triage, edstays, and diagnosis, as they only cover part of the patient's stay. The content of ``radiology'' will be set to the content of the section ``Imaging'' if the ``Imaging'' section is empty in the discharge letter. We then calculated the average ranking of the metric score for each section relative to the target sections, using the provided evaluation metrics, including BLEU-4 \citep{papineni2002bleu}, ROUGE-1/2/L \citep{lin2004rouge}, BERTScore \citep{zhang2019bertscore}, Meteor \citep{banerjee2005meteor}, AlignScore \citep{zha2023alignscore}, and MEDCON \citep{yim2023aci}. Each section was compared to the target sections, ``Brief Hospital Course'' (BHC) and ``Discharge Instructions'' (DI), with higher-ranking sections being more related to the target sections. Table \ref{tab:ranking} shows that ``History of Present Illness'' is most related to the BHC section, followed by imaging results, physical exams, past medical history, and diagnoses. BHC is most related to DI, followed by sections related to BHC.

\begin{table}[hbp]
\centering
\begin{tabular}{@{}lll@{}}
\toprule
\multicolumn{1}{c}{\textbf{Section}} & \multicolumn{1}{c}{\textbf{BHC}} & \multicolumn{1}{c}{\textbf{DI}} \\ \midrule
Patient Admissions         & 13 & 21 \\
Transfer Summary           & 15 & 23 \\
Diagnoses                  & 5  & 4  \\
Service                    & 11 & 12 \\
Allergies                  & 14 & 22 \\
Attending                  & 17 & 24 \\
Chief Complaint            & 8  & 11 \\
Major Surgical Procedure   & 9  & 17 \\
History of Present Illness & 1  & 2  \\
Review of System           & 10 & 15 \\
Past Medical History       & 4  & 9  \\
Social History             & 16 & 25 \\
Family History             & 12 & 16 \\
Physical Exam              & 3  & 5  \\
Pertinent Results          & 7  & 18 \\
Imaging and Studies        & 2  & 3  \\
Brief hospital course      &    & 1  \\
Admission Medications      &    & 10 \\
Discharge Medications      &    & 7  \\
Discharge Disposition      &    & 14 \\
Discharge Diagnoses        &    & 6  \\
Discharge Condition        &    & 8  \\
Followup Instructions      &    & 13 \\
Provider                   &    & 19 \\
Code Status                &    & 20 \\ \bottomrule
\end{tabular}
\caption{The ranking of different sections' relation to BHC/DI by averaging all the evaluation metrics provided by this challenge. We aggregated the patient's admission info, including gender, race, age (calculated), diagnosis, and transfer history from other MIMIC-IV tables.}
\label{tab:ranking}
\end{table}

Based on the ranking in Table \ref{tab:ranking} and the length of each section, we selected ``History of Present Illness'', ``Imaging and Studies'', ``Past Medical History'', ``Patient Admissions'', and ``Chief Complaint'' as the context for the BHC section. We used the generated BHC, ``Discharge Medications'', ``Discharge Disposition'', ``Discharge Diagnoses'', ``Discharge Condition'', and ``Followup Instructions'' for DI section. Other sections related to DI were excluded because they are also related to BHC. We truncated each section to the 95th percentile of its total length to remove outliers and potential segmentation errors.

\subsection{Retrieval for the Target Section Word Count}

\begin{figure*}[hbpt]
  \includegraphics[width=1\linewidth]{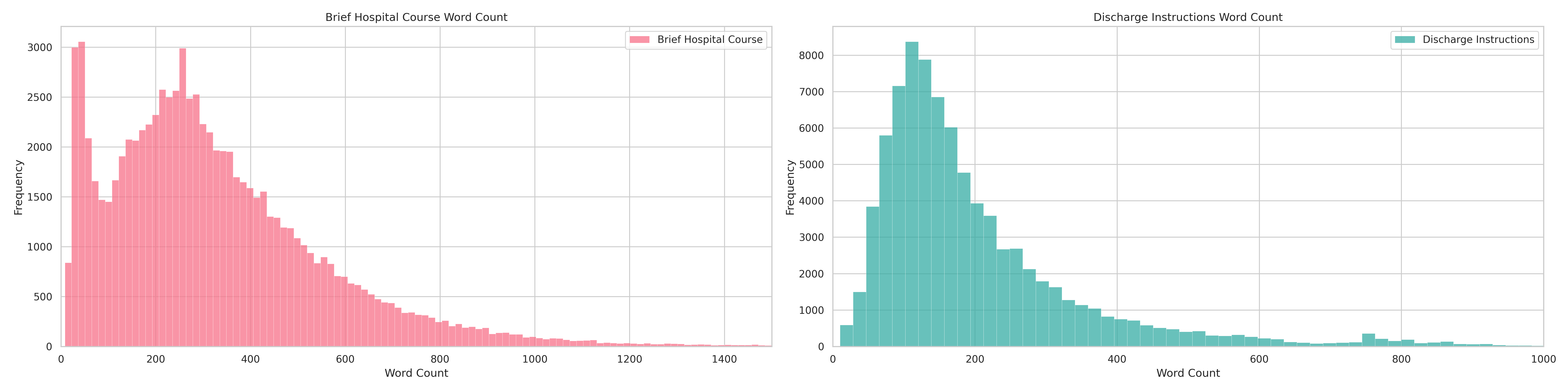}
  \caption{The target section word count distribution. Both BHC and DI have right-skewed distributions. BHC has two peaks, one below 100 words and one around 250 words.}
  \label{fig:target_count}
\end{figure*}

Understanding the target section's word count is beneficial for generating the appropriate amount of text, thereby improving the evaluation metrics for this challenge. Figure \ref{fig:target_count} shows the word count distribution for the target sections in the training dataset. Both target sections have right-skewed distributions, and BHC also has a peak for word counts under 100. We hypothesize that patients with similar backgrounds may have similar target sections. These retrieved target sections from patients with similar backgrounds can be used as a starting point, providing a template or word count for further refinement. We selected ``Chief Complaint'', ``Diagnoses'', and ``History of Present Illness'' as inputs for retrieving the BHC section. We added ``Admission medications'', ``Discharge Medications'', ``Discharge Disposition'', ``Discharge Diagnoses'', and ``Discharge Condition'' for retrieving the DI section. We used the \href{https://huggingface.co/sentence-transformers/all-MiniLM-L6-v2}{``sentence-transformers/all-MiniLM-L6-v2''} model to create embeddings of the context information for each training dataset entry and FAISS for similarity search. The word count from the first retrieved document's target section was used in the prompt to LLM for the generation. We compared this word count selection strategy to using a fixed word count, and the results are presented in Section \ref{sec:results}.

\subsection{Target Section Structure Template and Prompt Creation}

The target word count distribution varies, and we inspected several randomly chosen examples of target sections with different word counts. We selected examples with word counts over 180 to accommodate most cases for BHC template construction. Examples with word counts between 100-300 were chosen for the DI template construction. The structure is in JSON format, with names and descriptions for each section.

The BHC structure template is:

\texttt{\begin{enumerate}
    \item Introduction: Brief introduction including patient demographics, significant past medical history, and reason for hospitalization.
    \item Active Issues: Details of the primary medical concerns addressed during the stay, including initial assessments and management actions.
    \item Chronic Issues (Optional): Management of known chronic conditions during the hospital stay.
    \item Transitional Issues (Optional): Specific follow-up actions recommended for post-discharge care.
    \item Additional Notes (Optional): Other pertinent information or considerations affecting patient care.
\end{enumerate}}

The template includes several optional sections not included in all the examples. The template will be fed to the prompt below as the ``structure'' variable. The prompt for BHC is:

\texttt{As a medical professional, you are tasked with drafting a ``Brief Hospital Course'' section for a discharge letter.
Utilize the structure from a brief hospital course example to guide your composition.
The goal is to write a new, coherent, brief hospital course for another patient based on the provided structured template.
The total word count for the brief hospital course should be \{words\} words.\\ \\
BHC Instructions:}
\texttt{\begin{enumerate}[label=\arabic*.]
    \item Follow the JSON template provided to structure the new brief hospital course. Each section should be filled according to the relevant patient information.
    \item Omit the optional sections if they are irrelevant to the patient's case.
    \item Omit the optional sections if the total word count is less than 100 words.
    \item Do not add a new section after Additional Notes.
    \item Use placeholders ``\_\_\_'' for any date, patient name, and location.
    \item Use appropriate medical terminology and concise language to ensure clarity and professionalism.
    \item Do not be wordy; be concise if possible.
    \item Do not include the word "optional" in the result if they are included. If they are not included, just omit those sections.
    \item Do not copy patient information verbatim; paraphrase and use the structure template to fit in the details.
    \item All the section headers must be from the template, not from the patient information.
    \item Do not fabricate details not present in the patient information.
    \item Use section headers for each major medical issue, starting with a hashtag \#, do not use * for section header.
    \item Use bullet points to highlight key actions, medication changes, or critical clinical decisions, starting with a hyphen -. Do not use * or +.
    \item Ensure that each major issue or condition has its own section header if there is enough content related to it, even if briefly mentioned.
    \item Write in a narrative style for each section, providing a detailed account of the patient's condition, treatment, and outcomes.
    \item Employ medical abbreviations and terminology appropriately to convey information efficiently.
    \item Start the output with ``Brief hospital course:''
\end{enumerate}}

\texttt{Example structure for the brief hospital course: \{structure\}.\\
Patient information: \{context\}. }

The template for DI is below. This is fed to the DI prompt as the ``structure'' variable.

\texttt{\begin{enumerate}
\item Greeting: ``Dear [Title] \_\_\_,'',
    ``HospitalExperience'': ``It was a pleasure taking care of you at \_\_\_.'',
\item AdmissionReason: {
       ``Title'': ``WHY WAS I ADMITTED TO THE HOSPITAL?'',
       ``Details'': ``[ReasonForAdmission]''
    },
\item  InHospitalActivities: {
      ``Title'': ``WHAT HAPPENED WHILE I WAS IN THE HOSPITAL?'',
      ``Details'': ``[ActivitiesDuringStay]''
    },
\item  DischargeAdvice: {
      ``Title'': ``WHAT SHOULD I DO WHEN I GO HOME?'',
      ``Instructions'': ``[PostDischargeInstructions]''
    },
\item Closing: ``We wish you the best!'',
    ``CareTeam'': ``Your \_\_\_ Team''
\end{enumerate}}

The prompt for DI is:

\texttt{You are tasked with drafting a ``Discharge Instructions'' section for a patient's discharge letter as a medical professional.
The instructions should succinctly summarize the key points of the patient's hospital stay and post-discharge care clearly and easily for the patient to follow.\\ \\
DI Instructions:}
\texttt{\begin{enumerate}[label=\arabic*.]
    \item Use the JSON template provided to structure the discharge instructions.
    \item Do not include explicit section headers in the final text, such as ``Greeting'' or ``Hospital Experience''.
    \item Do not include any placeholder such as ``[]'' in the result.
    \item Include the title in the template.
    \item Integrate medication information narratively, mentioning specific medications only when discussing their relevance to the patient’s ongoing care and follow-up instructions.
    \item Do not list medications; describe how they contribute to the patient’s treatment plan.
    \item The total word count should be around \{words\} words, focusing on essential instructions relevant to the patient's care.
    \item Use ``\_\_\_'' to anonymize any date, patient name, and location.
    \item Clearly specify any medication changes, follow-up appointments, and additional care instructions using placeholders where specific details are to be inserted.
    \item Employ a professional yet empathetic tone to ensure clarity and approachability.
    \item Integrate medical terminology appropriately, ensuring it is understandable to a layperson.
    \item Start the output with a polite greeting and conclude with well-wishes or a thank you message.
\end{enumerate}}

\texttt{Example structure for the discharge instructions: \{structure\}.\\
Patient information: \{context\}.}

\section{Results}\label{sec:results}

The Llama3 model was downloaded from the Ollama model repository with the model ID \href{https://ollama.com/library/llama3:8b-instruct-q8_0}{``llama3:8b-instruct-q8\_0''}. We utilized the LangChain framework for retrieval, template building, and model calling. All experiments were conducted on a T4 GPU with 16GB memory, using the Microsoft Azure platform's ``Standard NC4 as T4 v3 (4 vCPUs, 28 GiB memory)'' configuration.

We compared several approaches:
\begin{enumerate}
    \item \textit{Baseline with Random Shuffling}: We shuffled the ``hadm\_id'' column, a unique identifier for each patient's discharge letter, assigning a random target section to each ``hadm\_id''. This random selection comes from the same distribution as the training data but without the actual content of the input text.
    \item \textit{Baseline with RAG Retrieval}: We used the retrieved target sections directly. This result can be similar to the target, but the details can differ from the real input.
    \item \textit{Fixed Target Word Count}: We set a fixed word count of 420 for BHC and 100-200 for DI in the prompt.
    \item \textit{Proposed Method}: Our method combines retrieved target word counts with a structured template.
\end{enumerate}

Table \ref{tab:result} presents the evaluation metrics from the Codabench platform \citep{xu-etal-2024-overview}, including BLEU-4 \citep{papineni2002bleu}, ROUGE-1/2/L \citep{lin2004rouge}, BERTScore \citep{zhang2019bertscore}, Meteor \citep{banerjee2005meteor}, AlignScore \citep{zha2023alignscore}, and MEDCON \citep{yim2023aci}. The random shuffle yielded the lowest scores across all metrics, indicating poor performance. Using the retrieved target section directly resulted in the highest BLEU score. The fixed word count approach achieved higher Align and MEDCON scores than the retrieved target section but had lower scores for other metrics. Our proposed method, which combines the retrieved word count and structured template, achieved the highest scores across all metrics except BLEU. The lower BLEU score for the proposed method is due to BLEU's heavy penalty for deviations from exact wording. In contrast, the higher ROUGE scores indicate our method effectively captures the essential content, even with varied wording. We also measured the generation time for each section. The average time to generate one BHC was 16.67 seconds, and one DI was 16 seconds.

\begin{table*}[htbp]
\centering
\begin{tabular}{@{}llllllllll@{}}
\toprule
 &
  \textbf{bleu} &
  \textbf{rouge1} &
  \textbf{rouge2} &
  \textbf{rougel} &
  \textbf{bertscore} &
  \textbf{meteor} &
  \textbf{align} &
  \textbf{medcon} &
  \textbf{overall} \\ \midrule
\textbf{random shuffle}        & 0.01           & 0.183 & 0.025 & 0.105 & 0.226 & 0.23  & 0.109 & 0.1   & 0.124 \\
\textbf{RAG retrieved  target} & \textbf{0.041} & 0.286 & 0.061 & 0.172 & 0.293 & 0.297 & 0.167 & 0.203 & 0.19  \\
\textbf{fixed target  word}    & 0.017          & 0.296 & 0.055 & 0.159 & 0.256 & 0.285 & 0.187 & 0.221 & 0.185 \\
\textbf{retrieved word  count} &
  0.024 &
  \textbf{0.377} &
  \textbf{0.106} &
  \textbf{0.205} &
  \textbf{0.3} &
  \textbf{0.332} &
  \textbf{0.174} &
  \textbf{0.254} &
  \textbf{0.221} \\ \bottomrule
\end{tabular}
\caption{The evaluation results from the Codabench platform. The random shuffle method yielded the lowest scores, while our final retrieval approach to determine the target word count achieved the highest scores across most metrics.}
\label{tab:result}
\end{table*}

\section{Unsuccessful Attempts}

\begin{figure}[hbpt]
  \includegraphics[width=\columnwidth]{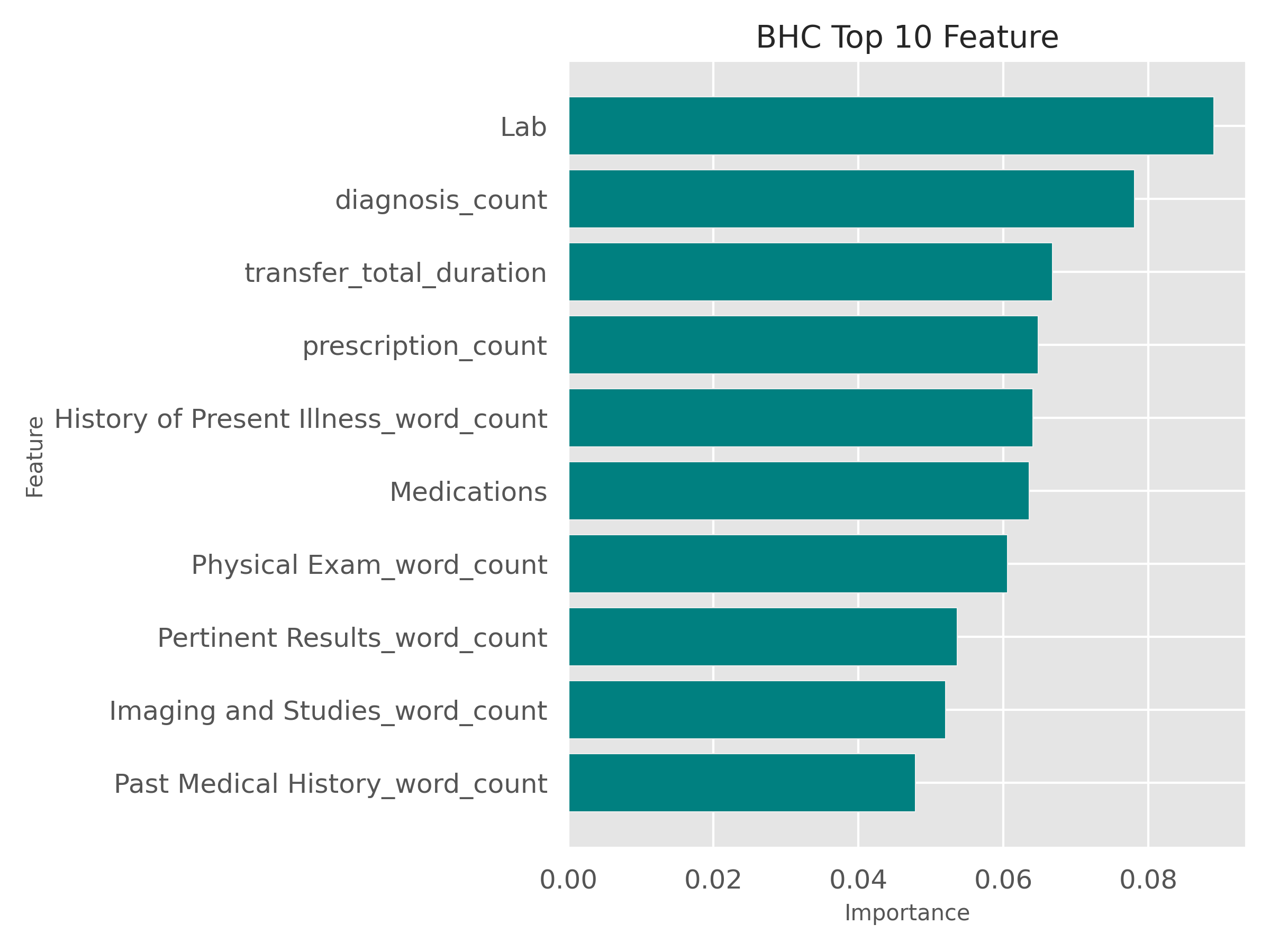}
  \caption{The top 10 features for the BHC classifier. WC: word count. The total number of lab tests, diagnosis, and total duration in the hospital are the top 3 features.}
  \label{fig:bhc_features}
\end{figure}

\begin{figure}[hbpt]
  \includegraphics[width=\columnwidth]{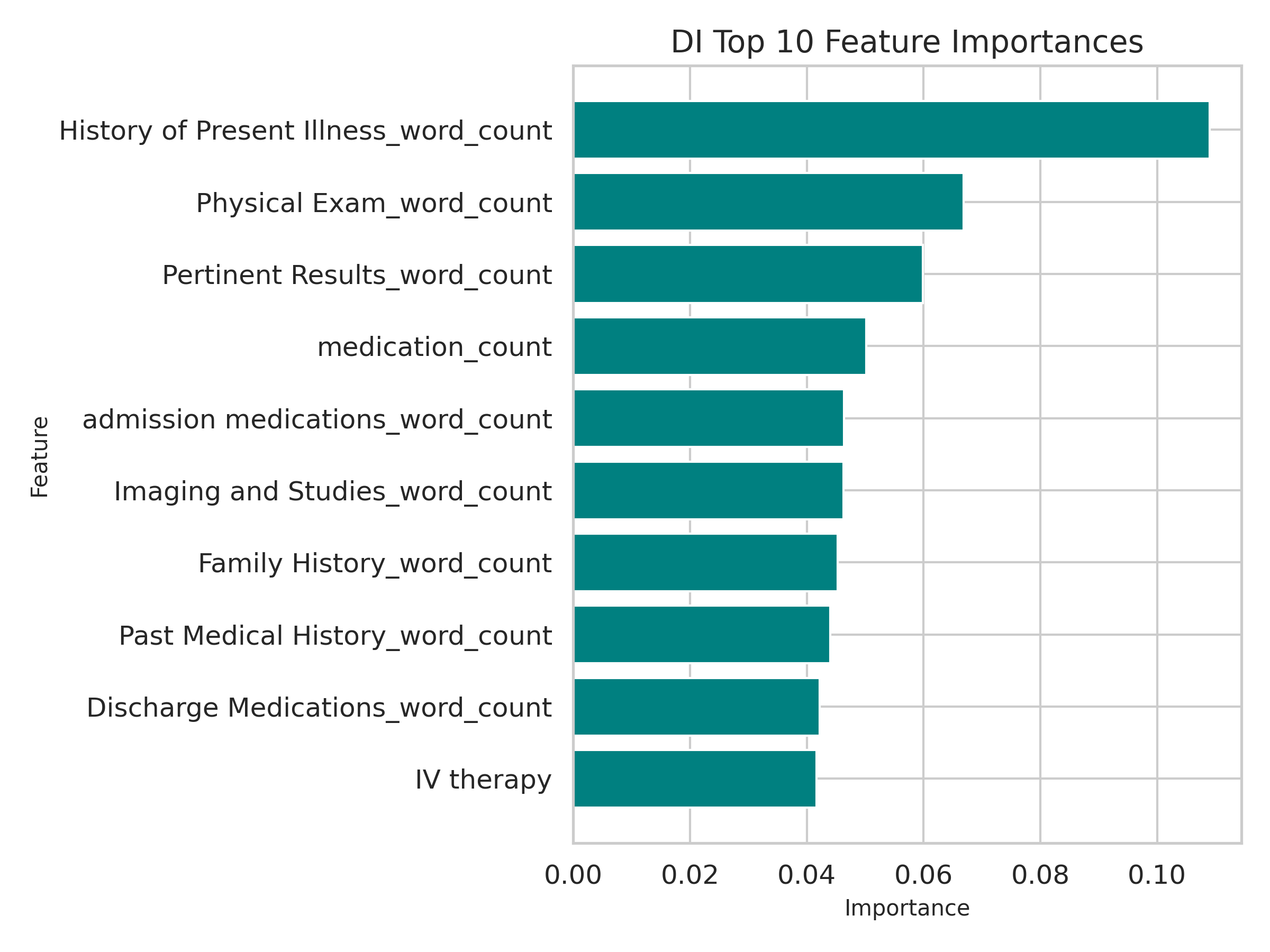}
  \caption{The top 10 features for the DI classifier. WC: word count. The word count of different segments is ranking high.}
  \label{fig:di_features}
\end{figure}

We also explored several alternative approaches for this task, but they yielded unsatisfactory results:

\begin{enumerate}
    \item \textit{Style Transfer Using Retrieved Target Section}: We asked the LLM to use the style of the retrieved target section to fit the patient context. However, the Llama3 8B model often used the target section directly, failing to infer the style and remove the original content. This could be due to the weaker reasoning ability of the 8B model compared to the 70B model with better reasoning ability.
    
    \item \textit{Two-Step Style Transfer}:
        \begin{enumerate}
            \item Firstly, extract a template from the target section.
            \item Secondly, fill in the patient content into the template (this step can also be split into several smaller steps).
        \end{enumerate}
        However, the extracted templates were not always reliable, and this method took twice as long as the curated template approach. Consequently, we opted to curate the templates rather than relying on the LLM manually.

    \item \textit{Predicting Target Section Word Count}: We tested several methods to predict the total word count of the target section, including fitting a random forest classifier by aggregating over 100 features from other MIMIC-IV tables and fitting log-normal distributions. These methods also proved inadequate. Table \ref{tab:bhc_classifier} shows the random forest classifier results for BHC with word count classes greater than 450, with an F1 score of 0.45. Figure \ref{fig:bhc_features} lists the top 10 features, including the number of lab tests, diagnoses, and total hospital duration. The classifier achieved an F1 score of 0.49 for word counts greater than 280 for the DI section, as shown in Table \ref{tab:di_classifier}, with different section word counts being the top features in Figure \ref{fig:di_features}.
\end{enumerate}

\begin{table}[htpb]
\centering
\resizebox{\columnwidth}{!}{%
\begin{tabular}{@{}lllll@{}}
\toprule
             & \textbf{precision} & \textbf{recall} & \textbf{f1-score} & \textbf{support} \\ \midrule
<450         & 0.818              & 0.926           & 0.869             & 18965            \\
>450         & 0.610              & 0.359           & 0.452             & 6087             \\\bottomrule
\end{tabular}%
}
\caption{BHC random forest classifier results for BHC word count above and below 450. The f1-score is 0.45 for the class with more than 450 words, which is not accurate enough.}
\label{tab:bhc_classifier}
\end{table}

\begin{table}[htb]
\centering
\resizebox{\columnwidth}{!}{%
\begin{tabular}{@{}cllll@{}}
\toprule
\multicolumn{1}{l}{} &
  \multicolumn{1}{c}{\textbf{precision}} &
  \multicolumn{1}{c}{\textbf{recall}} &
  \multicolumn{1}{c}{\textbf{f1-score}} &
  \multicolumn{1}{c}{\textbf{support}} \\ \midrule
<280         & 0.864 & 0.964 & 0.911 & 20143 \\
>280         & 0.716 & 0.377 & 0.494 & 4909  \\\bottomrule
\end{tabular}%
}
\caption{DI random forest classifier result for DI word count above and below 280. The f1-score is 0.49 for the class with more than 280 words, which is not accurate enough.}
\label{tab:di_classifier}
\end{table}

\section{Conclusion}

In this paper, we present a resource-friendly approach to automating the generation of the ``Brief Hospital Course'' and ``Discharge Instructions'' sections in discharge letters using the Llama3 8B quantized model. Our zero-shot template-based method and Retrieval-Augmented Generation produce high-quality, contextually appropriate summaries. However, we observe a lower BLEU score due to the different wording between the method’s result and the target sections. Ensuring the reliability and accuracy of generated content remains a significant challenge. Future work will focus on enhancing model reasoning capabilities, improving dynamic template extraction, and integrating robust validation mechanisms to verify medical accuracy. The code for this work is shared on \url{https://github.com/ruiguo-bio/discharge_me}, covering aggregating additional tables, segmentation of the discharge letters, RAG for the two target sections, and the random forest classifier for the target section words prediction.

\section{Limitations and Future Work}
\begin{enumerate}[label=\arabic*.]
    \item We would like to perform a more thorough evaluation to ensure that the model’s generated content is clinically relevant and does not include false or harmful information. This evaluation could be extended to understanding the strengths and weaknesses of language models for the challenge task.
    \item We create a template by sampling target sections with word counts close to the median. However, the length and structure of real target sections can vary significantly from our template. Our approach could be improved by predicting the target word count more precisely or by sampling different templates depending on the word count.
    \item We would like to test a wider range of language models and thoroughly compare different methods of providing relevant context to the language model, including different methods of Retrieval-Augmented Generation (RAG) and prompt engineering.
   
\end{enumerate}

\section{Ethical Statement}

All the data used in the experiments are downloaded from the PhysioNet after completing the required CITI training and credentialing process. Beyond the general potential ethical considerations of using LLMs to automatically process and generate clinical text (including bias, fairness, transparency and accountability), there are no specific ethical issues raised by the particular methodologies or data presented in this research.

\bibliography{acl_latex}

\end{document}